\documentclass[10pt,letterpaper]{article}
\usepackage[top=0.85in,left=1.75in,footskip=0.75in,marginparwidth=2in]{geometry}
\usepackage{tabu}
\usepackage[utf8]{inputenc}

\usepackage{cite}

\usepackage{nameref,hyperref}

\usepackage{microtype}
\DisableLigatures[f]{encoding = *, family = * }

\usepackage{changepage}

\usepackage[aboveskip=1pt,labelfont=bf,labelsep=period,singlelinecheck=off]{caption}

\makeatletter
\renewcommand{\@biblabel}[1]{\quad#1.}
\makeatother

\usepackage{lastpage,fancyhdr,graphicx}
\usepackage{epstopdf}
\fancyheadoffset[L]{2.25in}
\fancyfootoffset[L]{2.25in}

\usepackage{color}

\definecolor{Gray}{gray}{.25}

\usepackage{graphicx}

\usepackage{sidecap}

\usepackage{wrapfig}
\usepackage[pscoord]{eso-pic}
\usepackage[fulladjust]{marginnote}
\reversemarginpar

\begin{document}
\vspace*{0.35in}

\begin{flushleft}
{\Large
\textbf\newline{An HTM based cortical algorithm for detection of seismic waves.}
}
\newline
\\
Ruggero Micheletto\textsuperscript{1,*},
Ahyi Kim\textsuperscript{1}
\\
\bigskip
\bf{1} Yokohama City University, Dept of Nanobiosystems
\\
\bigskip
* ruggero@yokohama-cu.ac.jp

\end{flushleft}

\section*{Abstract}
Recognizing seismic waves immediately is very important for the realization of efficient disaster prevention. Generally these systems consist of a network of seismic detectors that send real time data to a central server. The server elaborates the data and attempts to recognize the first signs of an earthquake. The current problem with this approach is that it is subject to false alarms. A critical trade-off exists between sensitivity of the system and error rate. To overcame this problems, an artificial neural network based intelligent learning systems can be used. However, conventional supervised ANN systems are difficult to train, CPU intensive and prone to false alarms. To surpass these problems, here we attempt to use a next-generation unsupervised cortical algorithm HTM\cite{Cui:2016}. This novel approach does not learn particular waveforms, but adapts to continuously fed data reaching the ability to discriminate between \textit{normality} (seismic sensor background noise in no-earthquake conditions) and \textit{anomaly} (sensor response to a jitter or an earthquake). Main goal of this study is test the ability of the HTM algorithm to be used to signal earthquakes automatically in a feasible disaster prevention system. We describe the methodology used and give the first qualitative assessments of the recognition ability of the system. Our preliminary results show that the cortical algorithm used is very robust to noise and that can successfully recognize synthetic earthquake-like signals efficiently and reliably.

\section*{Introduction}
This is a brief report on the setup of a Hierarchical Temporal Memory (HTM) cortical model for real-time detection of seismic waves. Generally, early detection of dangerous seismic phenomena can be done by hybrid methodology, like on-line analysis of amplitude, duration and spectrum of the vibration\cite{rex:1978}. However, this approach is subject to a high rate of misses and false alarm. Errors are caused by human induced spurious signals due to automotive traffic, local mechanical disturbances or other phenomena. To reduce them it is necessary to distinguish between these unwanted signals and real seismic waves embedded in noise. This can be done with an intelligent software able to learn the shape of a natural seismic and recognize it from a background of noise and human-induced disturbances. For this purpose, artificial neural networks have been proposed. These networks are able, if properly trained by trains of data with examples of real seismic waves and noises, to classify wave shapes and deduce if they are dangerous seismic waves or innocuous artificial disturbances. However, training of these networks is difficult and precision of classification is still low\cite{dai:1995,akra:2016,adel:2009}.
\begin{figure}
\includegraphics[width=\textwidth]{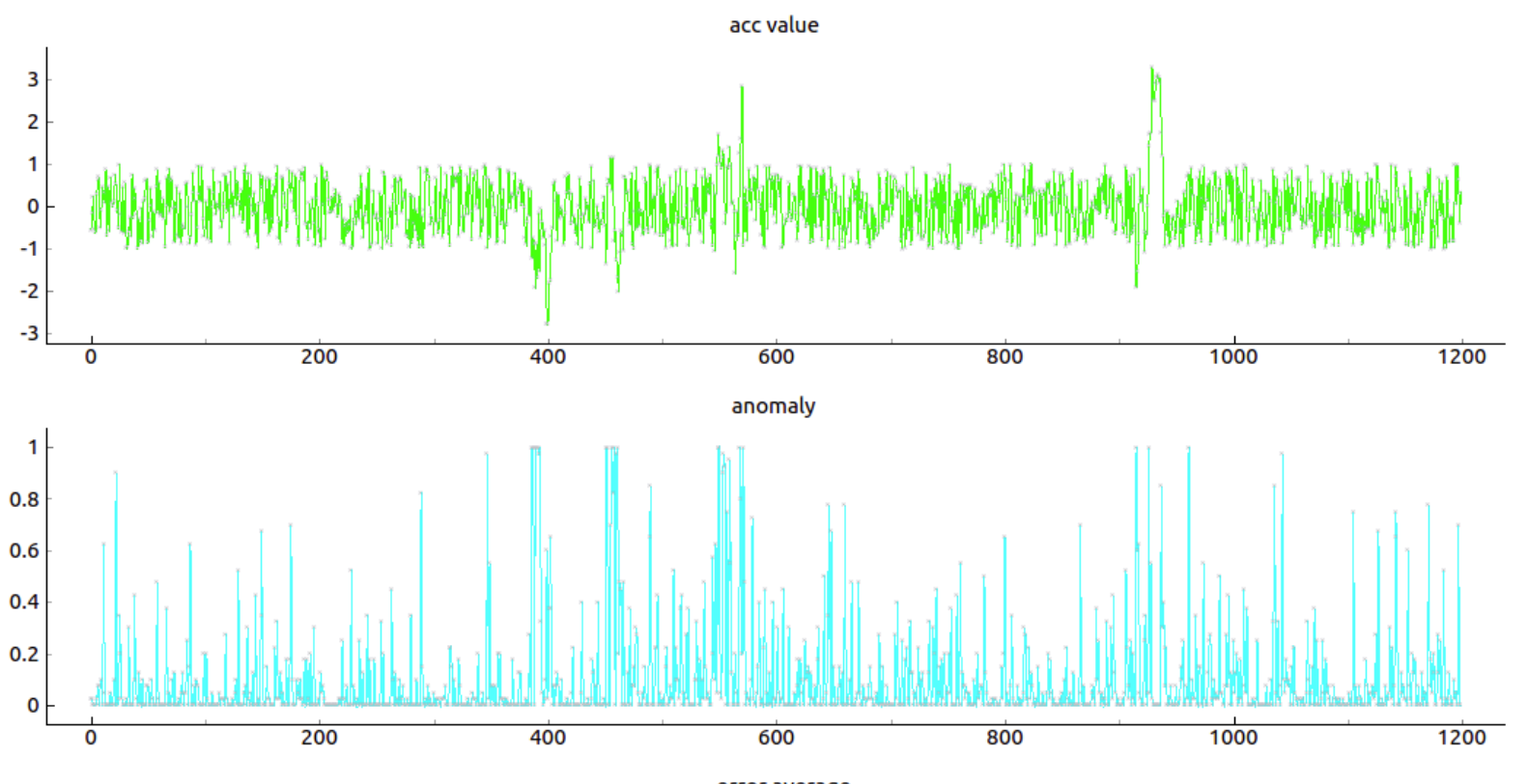}
\caption{Top panel: the synthetic seismic waveform fed to the HTM model. This is made by a uniform random acceleration signal between -1 and 1 and sporadic \textit{seismic waves}. These are synthesized by equation \ref{eq:genwave} with a probability $p=0.005$, that corresponds, on average, once every 200 time steps. Bottom panel: the anomaly score output of the model. This plot shows the first 1200 points given to the network. In this initial phase, the cortical network is not adapted to the signal and seems to identify the instrumental noise as signal anomaly. The score keeps significantly higher than zero, with variable values all along the plot. Overall, the network looks incapable to reliably distinguish the simulated seismic waves (for example at about $t=400$, $t=600$ and $t=750$) from the simulated background noise. Time $t$ is an arbitrary sample interval here, in a real experiment this interval will be the integration time of the seismic sensor.} 
\label{fig:fewMin} 
\end{figure} 
\medskip
To overcame these problems, we test here an anomaly-detection unsupervised algorithm based on biological plausible cortical structure. These models are developed on a new theoretical framework that construct a hierarchical temporal memory system able to learn a sequence of data that is fed continuously to the input. These HTM systems perform similarly to other sequence learning algorithms like autoregressive moving averages, feedforward neural networks and recurrent neural networks, but have the advantage to be ready for continuous online learning and are inherently specialized for anomaly detection. \par The seismic sensors will feed continuously data to the network. The data are synthetic, generated by an algorithm in order to represent a combination of instrumental noise and human generated spurious signals (people walking in the room nearby the sensor, slamming doors, traffic nearby etcetera). The cortical algorithm is fed continuously with this data that will be learned as the \textit{normality} by the system. Our hope is that deviation from this normality (an earthquake) will result in an anomaly signal by the HTM. This approach has the advantage that we do not need training, the network will learn by itself what is normal and what it is an anomaly in a unsupervised way. Since we plan to realize a network of sensors placed in disparate places, an HTM network will learn specific local disturbances, whereas standard supervised methods may need to be re-calibrated for sensors located in different places. An HTM network has also the ability to predict future evolution of the signal, it is robust to sensor noise, has fault tolerance and exhibit good performance without the need for task-specific tuning\cite{Cui:2016}.
\begin{figure}[ht] 
\includegraphics[width=\textwidth]{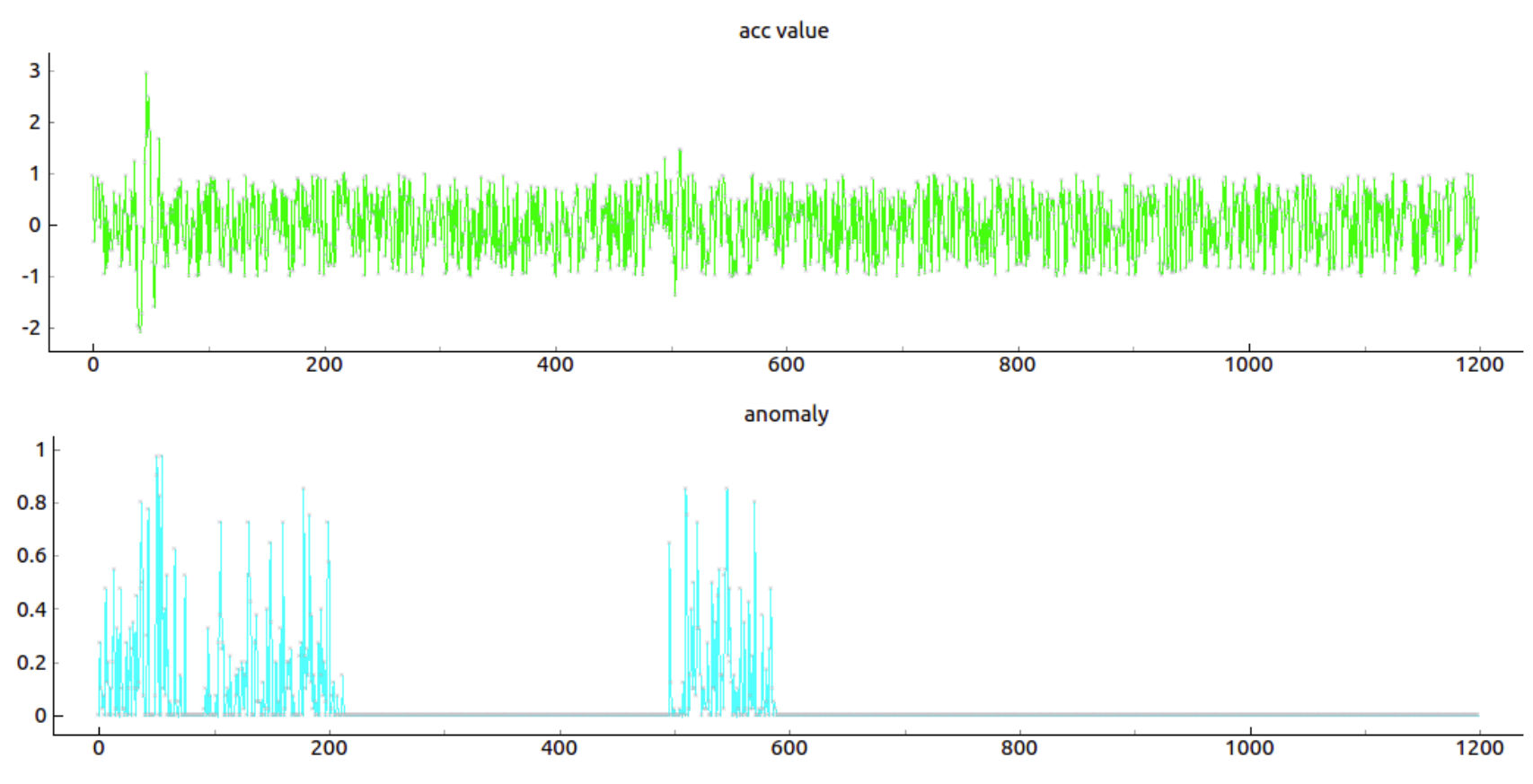}
\caption{After about $t=450.000$ time steps (plot shows a window of 1200 of them), the network lowers its anomaly output, and gains an impressive robustness to noise. The anomaly score drops cleanly to zero when random noise is constantly fed to the model. This results in distinct identification of the simulated seismic waves visible in the signal of the top panel at $t\approx~50$. Moreover, small waveforms, barely distinguishable by eye, give rise to net anomaly score around $t=500$ in the bottom panel.}
\label{fig:360x1200} 
\end{figure}
\section*{Method}
For convenience seismic waves were synthesized by a simple algorithm. This gives us the advantage to experiment with the HTM network with calibrated and optimal data very similar to the expected real case scenario signals. We didn't use real sensors data in this case, because those should be placed in a noisy laboratory where all sort of activities are on going. This in practice will limit the amount of usable data for our tests. \par The algorithm basically simulate both instrumental noise and small jitters that represent human noises and vibrations in real experimental environments. The instrumental noise is simulated with a simple uniform distribution of numbers withing the interval ${-1, 1}$. These values are representing normalized accelerations. Jitters are instead occurring with a probability of $p=0.005$ (on average once every 200 points) and are generate summing up several random sine waves. Overall the signal is obtained with this formula:
\begin{eqnarray}
A_c(t)=\Sigma_n{a sin(2\pi f_n t)} + \epsilon
\label{eq:genwave}
\end{eqnarray}
where $f_n$ are ten frequencies chosen at random in the interval $f_{min}=0.01$ and $f_{max}=0.1$. The parameter $\epsilon$ is the uniformly distributed instrumental noise mentioned above. The jitter duration is fixed to 25 time units and its amplitude $a$ is a random number between 0 to five. 
\par
The calculations are done by a 64 bits Linux (Ubuntu) machine with 6 Gb Ram and 4 threads CPU. Numenta NuPIC was installed on the system\cite{numenta} to setup the HTM cortical network. NuPIC is a package that implements the HTM network structure, the user can manipulate the parameters that regulate its functioning.

To determine the best configuration for our HTM network we did a parameter search through a \textit{swarming} process\cite{swarming}. The final relevant parameters of our cortical network are the one in table \ref{tab:pars}. Notice that those numbers are given in NuPic's conventional order and naming. To understand exactly the meaning of the table, readers should refer to HTM literature (see provided literature about HTM and NuPic\cite{Cui:2016,numenta}). Usage of NuPic HTM cortical algorithms will generate parameters that can compare directly to this table.   
\begin{table}[!ht]
\centering
\caption{{\bf Cortical Algorithm Parameters} The parameters used by the HTM cortical algorithm. This table is given as a guideline. Sp and tp parameters are give in the HTM conventional order. Readers should learn details about the functioning of HTM algorithm to understand fully the meaning of the values.}
\begin{tabular}{m{4cm} m{4cm} m{4cm}}
\hline
\multicolumn{3}{|c|}{\bf HTM parameters} \\ \hline
alpha = 0.009340 & SDRClassifierRegion & steps = 1   \\ 
Verbosity = 0 & inference = anomaly & \\ 
\hline
\multicolumn{3}{|c|}{\bf sensor parameters} \\ \hline
clip = True & max = 2.0 & min = -2.0  \\ 
 n= 118 & type = scalarEncoder & w = 21   \\ 
\end{tabular}
\begin{tabular}{ m{12cm}} 
\hline
\multicolumn{1}{|c|}{\bf sp parameters} \\ \hline
0.0, 2048, 1, 0, 40, 0.8, 1956, 0, cpp, 0.05, 0.1, 0.1 \\ 
\hline
\multicolumn{1}{|c|}{\bf tp parameters} \\ 
\hline
12, 32, 2048, 0, 0.21, 2048, 0, 128, 32, 9, 20, normal, 1, 0.1, 0.1, 1960, cpp, 0 \\ 
\end{tabular}
\label{tab:pars}
\end{table}

\begin{figure}[ht] 
\includegraphics[width=\textwidth]{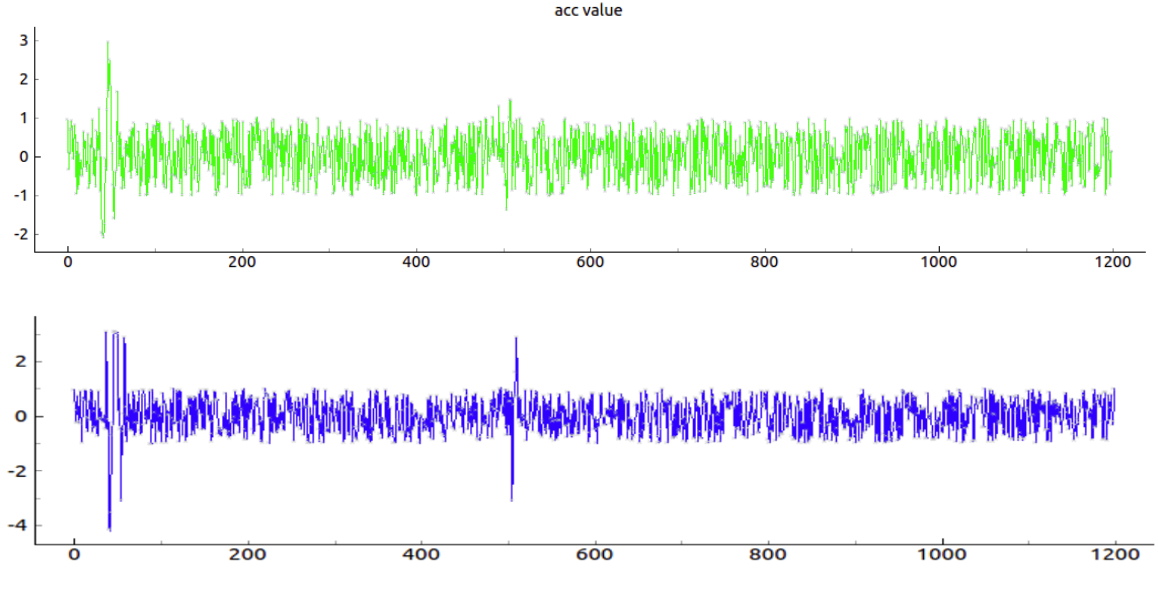}
\caption{The acceleration signal (top) and the corresponding HTM model predicted value. The HTM algorithm predictions are made one step ahead and are similar to input values.}
\label{fig:prediction} 
\end{figure}

\section{Results and Discussion}
Once the network is correctly setup with the parameters in Tab (\ref{tab:pars}), the model is run. We process an infinite loop in which synthetic seismic noise and waveforms are generated by expression \ref{eq:genwave}, each generated value at each time-step is fed to the model that uses the  \textit{multiStepBestPredictions} method to attempt prediction of the next acceleration value. Also the model outputs an \textit{anomalyScore} value ranging from 0 to one. This value represents how the model \textit{feels} about the current signal behavior, if it is considered highly anomalous, the score will be high, otherwise will be low. The anomaly score ranges from zero to one.\par   
In figure fig. \ref{fig:fewMin} are shown the first few thousands time steps of the model simulation. Each time steps corresponds to an arbitrary unity of time, in an experiment run with a real seismic sensor this time-step will correspond to the device integration time.

\par
Clearly the model is not able at this stage to discriminate between noise and anomalous seismic waves. Those are represented in fig. \ref{fig:fewMin} top panel at about $t=600$ and $t=750$. The cortical algorithm considers random noise as an anomaly and other waveforms too. This behavior makes sense since there is nothing more unpredictable and \textit{anomalous} than a random variable and other waveforms are new and unpredictable as well. \par
However, if we let the system run for many thousands cycles, the model begins to adapt to the continuous random noise, and eventually the anomaly score drops stably to zero each time a sequence of noise arrives. This shows how the HTM cortical algorithm has adapted to the random signal continuously fed to it. This somewhat compares to human or animal behavior when external disturbances are ignored if they are repeated regularly. In figure \ref{fig:360x1200} lower panel we see the response of the algorithm to the random seismic signal after about half a million time steps.  As shown in the lower panel, the model anomaly score to pure noise keeps nearly zero in a robust and reproducible manner. Interestingly, the simulated seismic waves instead are recognized as anomalous. The seismic signals visible in the top panel are correctly identified with higher anomaly scores. These fluctuations have their characteristics and are much less common than the random noise, nevertheless, after the adaptation period, the model becomes able to distinguish successfully between the two. Noticeably the duration of anomalous response is slightly delayed compared to the seismic waves. For example the short perturbation at about $t=500$ in the top panel corresponds to a longer anomaly score response. This particular waveform is very small and barely recognizable by eye, nevertheless the cortical algorithm was able to distinguish it from the background noise. This is an outstanding behavior that may prove to be useful for the realization of a disaster prevention system.
\par

Simultaneously to the anomaly scores, the model outputs a prediction of the next acceleration value, basing itself to previous accelerations sequences. In figure \ref{fig:prediction} we show a plot of the acceleration signal and the corresponding predictions.
Interestingly the prediction signal looks very similar to the actual waveform, but it precedes the waveform output by one steps. The system is able to somehow anticipate future seismic accelerations, this could be a remarkable feature in a HTM based disaster prevention system.\par
\begin{figure}[ht] 
\includegraphics[width=\textwidth]{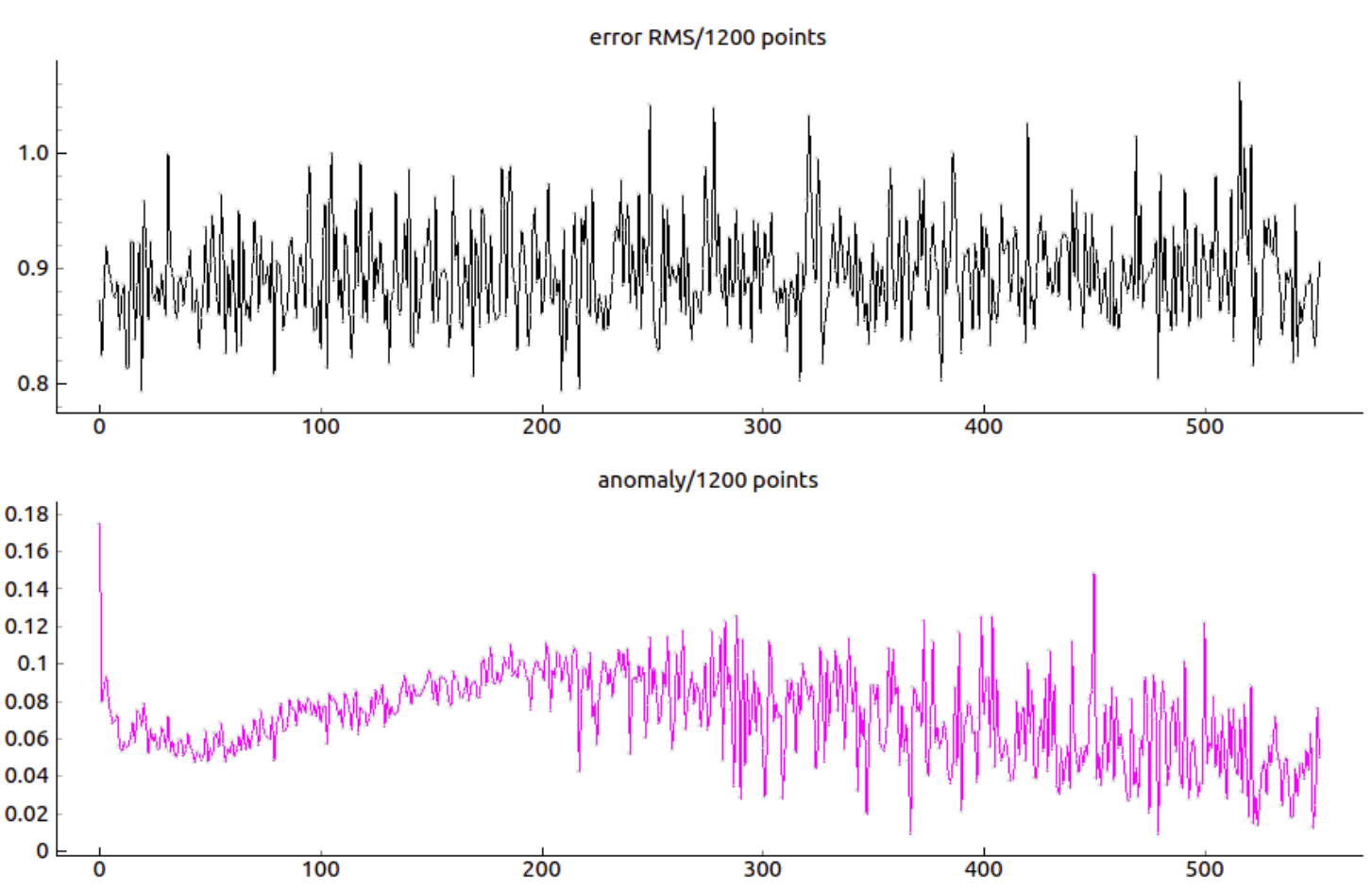}
\caption{In the top panel each points represents the RMS average value for 1200 points of simulation. The value is simply the average difference between the predicted value and the actual acceleration value for 1200 points (so total simulation was running for about 600x1200 time steps). The average error seems to remain stable all along the simulation. The bottom panel shows the averaged anomaly score calculated in the same fashion. The value is dropping at the very beginning, then rising again and the decreasing with high volatility. This particular shape of the average anomaly was reproduced in different experiments with different random seeds.}
\label{fig:ErrAno} 
\end{figure}

To evaluate the model's ability of prediction we calculated the error and average it for every time window (1200 time step points). The graph in figure \ref{fig:ErrAno} on the top shows this value over a span of over one million time steps (horizontal axis are averaged steps, each of them is 1200 simulation time-steps). The bottom panel of this figure represents the anomaly score averaged on 1200 points, and a characteristic rise and fall of this value is noticeable. This graph represents the \textit{learning curve} of the cortical algorithm. The first steep drop denotes the adaptation to the sensor background noise. After about 12.000 time steps the anomaly average is reduced of about 50\% the initial value. However, it takes about 250.000 time steps before the adaptation is complete. After full adaptation, anomaly stays stably to zero, when no seismic waveforms are in input. Subsequently the average anomaly seems to drop, but gets much more volatile, indicating the stronger influence of the random seismic waveform (see figure \ref{fig:ErrAno} bottom panel after $t\approx~200$).

\section{Conclusions}
We have evaluated for the first time how a cortical HTM algorithm can be used to recognize anomalies in seismic signals. We adopted the recent NuPic HTM model \cite{numenta,Cui:2016} and tested its performances on a simulated earthquake prediction experiment. In our setup the HTM model is fed continuously data from a seismic device. For the most of the time, these data are instrument background noises, however at a defined probability, a seismic waveform is added to the noise and we want to evaluate the HTM model ability to distinguish this from the background. We found that this system adapt very quickly to the random fluctuations. After an initial transition time of about 200 thousands time steps, the HTM cortical algorithm was able to consider the seismic sensor background noise as \textit{normality}, lowering its \textit{anomaly score} to nearly zero. Waveforms instead were recognized reliably with repeated spikes of high anomaly score. \par Our tests indicate that the HTM system it is robust to noise, and able to recognize efficiently small anomalies hidden in the signal. Our study is still qualitative and more investigations are necessary to characterize fully the performance of the algorithm as a feasible earthquake detector. However, on the basis of these results, we feel to speculate that an HTM setup can help to the realization of robust earthquake detection algorithms and contribute to intelligent anti-disaster programs.



\setcounter{figure}{0}
\renewcommand{\thefigure}{S\arabic{figure}}



\section*{Acknowledgments}
We thank Kahoko Takahashi for the help with citations and for reading this manuscript.




\end{document}